\documentclass[11pt]{article}

\usepackage[preprint]{acl}

\usepackage{times}
\usepackage{latexsym}

\usepackage[T1]{fontenc}

\usepackage[utf8]{inputenc}

\usepackage{microtype}

\usepackage{inconsolata}

\usepackage{graphicx}

\usepackage{amssymb}
\usepackage{amsmath}
\usepackage{pifont}
\usepackage{multirow}
\usepackage[vlined,ruled,boxed,commentsnumbered,linesnumbered]{algorithm2e}
\usepackage{listings}
\usepackage{tcolorbox}
\definecolor{dkgreen}{rgb}{0,0.6,0}
\definecolor{gray}{rgb}{0.5,0.5,0.5}
\definecolor{mauve}{rgb}{0.58,0,0.82}
\title{AgentGC: Evolutionary Learning-based Lossless Compression for \\Genomics Data with LLM-driven Multiple Agent}

\author{
  \textbf{Hui Sun\textsuperscript{1,2}},
  \textbf{Yanfeng Ding\textsuperscript{1}},
  \textbf{Huidong Ma\textsuperscript{1,2}},
  \textbf{Chang Xu\textsuperscript{3}},
  \textbf{Keyan Jin\textsuperscript{4}}, \\
  \textbf{Lizheng Zu\textsuperscript{2}},
  \textbf{Cheng Zhong\textsuperscript{5}},
  \textbf{Xiaoguang Liu\textsuperscript{1\thanks{Corresponding author}}},
  \textbf{Gang Wang\textsuperscript{1$^{*}$}},
  \textbf{Wentong Cai\textsuperscript{2}},
\\
  \textsuperscript{1} Nankai University,
  \textsuperscript{2} Nanyang Technology University, 
  \textsuperscript{3} Microsoft Research Asia, \\ 
  \textsuperscript{4} Macao Polytechnic University , 
  \textsuperscript{5} Guangxi University, 
\\
}
\usepackage{titlesec}

\begin{document}
\maketitle
\begin{abstract}
Lossless compression has made significant advancements in Genomics Data (GD) storage, sharing and management. Current learning-based methods are non-evolvable with problems of low-level compression modeling, limited adaptability, and user-unfriendly interface. To this end, we propose AgentGC~\footnote{https://anonymous.4open.science/r/AgentGC-C0F7.}, the first evolutionary  \underline{\textbf{Agent}}-based \underline{\textbf{G}}D \underline{\textbf{C}}ompressor, consisting of 3 layers with multi-agent named Leader and Worker. Specifically, the 1) User layer provides a user-friendly interface via Leader combined with LLM; 2) Cognitive layer, driven by the Leader, integrates LLM to consider joint optimization of algorithm-dataset-system, addressing the issues of low-level modeling and limited adaptability; and 3) Compression layer, headed by Worker, performs compression \& decompression via a automated multi-knowledge learning-based compression framework. On top of AgentGC, we design 3 modes to support diverse scenarios: CP for compression-ratio priority, TP for throughput priority, and BM for balanced mode.
{Compared with 14 baselines on 9 datasets, the average compression ratios gains are 16.66\%, 16.11\%, and 16.33\% , the throughput gains are 4.73$\times$, 9.23$\times$, and 9.15$\times$, respectively.}
\end{abstract}

\section{Introduction}
Genomic Data (GD), consisted of \{A, C, G, T\}, plays a crucial role in precision medicine, virus tracing, and new drug development~\cite{PMKLC,CGPU-F3SR}. Recently, with the advancement of AI-driven bio-informatics Large Language Models (LLMs) such as AlphaFold2~\cite{AlphaFold2}, GENERanno~\cite{GENERanno}, and GenomeOcean~\cite{GenomeOcean}, along with third-generation genome sequencing technologies, GD has entered an era of massive big data~\cite{ma2023ricme, NNLCB}.
For example, as of January 2026, the China National GeneBank Sequence Archive has backed up over 18,441.33 TB of big GD~\footnote{https://db.cngb.org/cnsa/statistic}. 
As a result, this presents significant challenges for GD storage and management. AI-based compression technology is an effective approach to alleviating this dilemma.

{\small
\begin{figure}[t]
  \centering
  \includegraphics[width=1.\columnwidth]{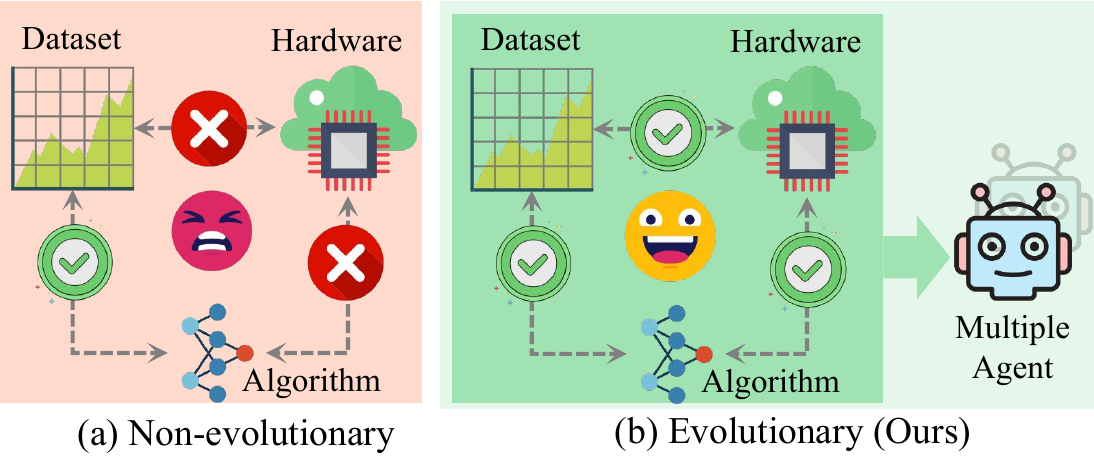}
  \caption{(a) The traditional non-evolutionary learning-based and (b) the proposed evolutionary scheme in AgentGC, which realizes collaborative modeling of ``Algorithm-Dataset-Hardware''.}
  \label{Fig_Classfiy}
\end{figure}
}

Although several multi-modal data compressors like MSDZip~\cite{MSDZip} and LLMZip~\cite{valmeekam2023llmzip} have been proposed, they fail to leverage the redundancy characteristics of GD for dedicate optimization, resulting in suboptimal compression ratio and throughput.
Fortunately, GD-dedicated Learning-based Compressors (GDLCs), such as DeepGeCo~\cite{DeepGeCo} and GenCoder
~\cite{GenCoder}, have filled this gap through refined data modeling.
However, in our investigation, existing GDLCs are non-evolutionary with fixed parameters and face the following challenges:

{\textbf{Low-level Compression Modeling}.} As illustrated in Figure~\ref{Fig_Classfiy}(a), traditional GDLCs are non-evolutionary: they only capture latent correlations between algorithms and datasets, reflecting a limited modeling granularity. In contrast, as shown in Figure~\ref{Fig_Classfiy}(b), we reframe compression as a system-level task by jointly considering the design of dataset, algorithm, and hardware memory.

{\small
\begin{figure}[t]
  \centering
  \includegraphics[width=1\columnwidth]{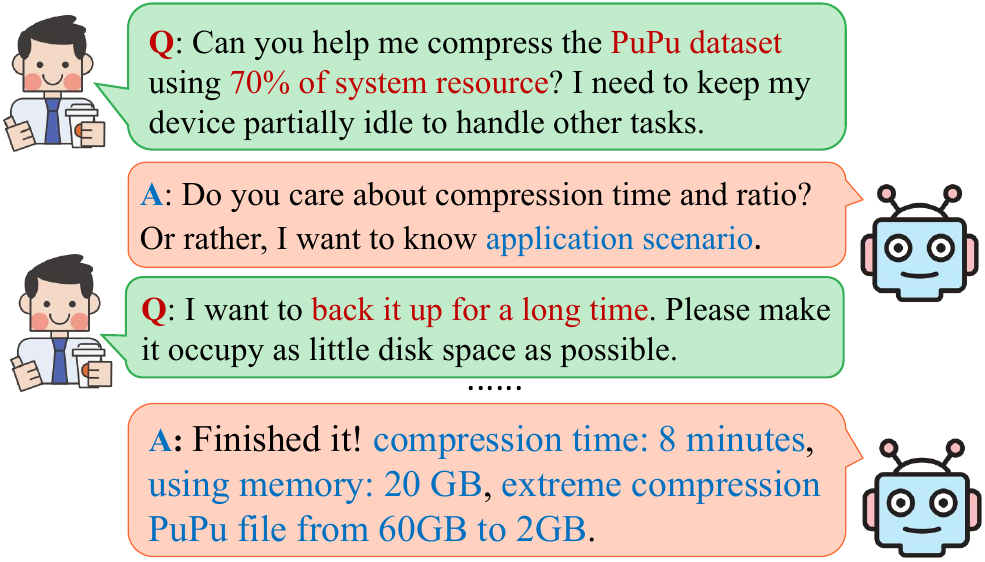}
  \caption{An example of genomic data lossless compression through interaction of LLM-based AgentGC.}
  \label{Fig_Demo}
\end{figure}
}
\textbf{Limited Adaptability}. Existing GD-dedicated compressors are typically tailored to single-use scenarios, limiting their effectiveness in complex environments. For example:  
1) Data persistence and backup applications require higher compression ratios;  
2) Real-time transmission demands faster compression and decompression speeds;  
3) Under memory constraints, balancing efficiency and performance becomes critical.

\textbf{User-unfriendly Interface}. The GDLCs involve the complex learning-based modeling, which introduces two disadvantages:
1) GDLCs encompass intricate model architectures, data distribution, parallelism, and additional configurations, making usage highly restricted for non-AI experts;
and 2) Joint tuning of multiple parameters in GDLCs are complex, and incorrect parameter configurations negatively impact overall performance.

To address these issues, as shown in Figure~\ref{Fig_Demo}, we propose AgentGC, the first novel evolutionary compression system for GD. The innovations and contributions of this paper are threefold:

1) We propose the first GD compression algorithm, AgentGC, which comprises a group of carefully designed leader agent and worker agent. Through dialogue-driven compression and automated LLM-based parameter tuning, AgentGC achieves more sophisticated modeling and provides a more user-friendly interface. 

2) Through prompt optimization and architectural design, we designed three modes for AgentGC, the a) CP for compression-ratio priority, b) TP for throughput priority, and c) BM for balanced mode, offering enhanced flexibility and adaptability across diverse scenarios.

3) We conducted a comprehensive comparison between AgentGC and 14 baselines across 9 datasets. AgentGC achieved up to a $75.456\%$ improvement in average compression ratio and a $18.995\times$ increase in throughput. AgentGC also demonstrated overall superiority in compression robustness and CPU \& GPU resource usage.

\section{Problem Definition}
Let $x=\{x_i\}_{i=0}^{m-1}\in \mathbb{R}^{1\times d_0}$ denote the inputed GD with length of $m$, the GDLC aim to generate a compact binary output $y=\{y_j\}_{j=0}^{n-1}\in \mathbb{R}^{d_1}$ with length of $n$. Here, $d_0$ and $d_1$ denote the length GD alphabet and the binary alphabet after compression, respectively. Generally, GDLC involves two stages:
1) Modeling, using an entropy function $\mathcal{F}$ to obtain the true probability distribution of $x$; and 2) Encoding, connecting an arithmetic encoder $\mathcal{E}$ to generate a compact representation. This process can be expressed as:
\begin{equation}
\label{defination1}
    y=\mathcal{E}(\mathcal{F}(x)).
\end{equation}
GDLC focus on modeling $\mathcal{E}(x)$ because it determines the compression efficiency of the encoder $\mathcal{E}$.
However, obtaining the real-world $\mathcal{F}(x)$ for GD is challenging~\cite{DZip,DeepGeCo}. Thus, GDLC fits $\mathcal{F}_{\theta}(x,v)$ by fixing a set of user-defined hyperparameters $v$ and utilizing neural network techniques, where $\theta$ denotes the learned parameters.
Therefore, the optimization objective of GDLC aims to minimize the difference $\mathcal{D}$ between $\mathcal{{F}}(x)$ and $\mathcal{F}_{\theta}(x,v)$:
\begin{equation}
\label{defination2}
    y\simeq \mathcal{E}\{ \mathcal{D}_{min.}(\mathcal{{F}}(x), \mathcal{F}_{\theta}(x,v))\}.
\end{equation}
The Cross-Entropy (CE) loss is widely used to learn this difference due to its close relationship with Shannon entropy~\cite{shannon1948mathematical}. Therefore, $\mathcal{D}_{min.}$ can be denoted as:
\begin{eqnarray}
\label{eq:ce}
\centering
\begin{aligned}
\mathcal{L}_{min.}\simeq&\sum_{i=c+1}^{m}\mathrm{CE}(\hat{f}_i,f_i)\\=&\sum_{i=c+1}^{m}\sum_{j=0}^{d_0-1}(\hat{f}_{i,j}log\frac{1}{f_{i,j}}).
\end{aligned}
\end{eqnarray}
Here, $c\in \mathbb{R}$ ($1\leq c\leq m-1$) is a user-defined \textbf{context length}. $\hat{f}_i\in\mathbb{R}^{1\times d_0}$ and $f_i\simeq \mathcal{F}_\theta^i\{(x_i|x_{i-t},...,x_{i-1}),v\}\in\mathbb{R}^{1\times d_0}$ denote the one-hot encoded ground truth and neural-network-predicated probability vector for \textbf{target symbol} $x_i$ using \textbf{historical symbols} $\{x_i|x_{i-t},...,x_{i-1}\}$.

In the enhanced GD compression system, the compressor automatically perceives and tunes the parameters $v$ through an LLM-driven agent and the user-provided prompt $t$. The preliminary definition is as:
\begin{equation}
\label{defination4}
    v\simeq  \mathrm{LLM}\varphi(t)
\end{equation}
Here, $\varphi$ denotes the weight parameters of LLM. The LLM-driven compression system can be formally defined as:
\begin{equation}
\label{defination5}
    \hat{y}\simeq\mathcal{P}(x|v)= \mathcal{P}\{\mathcal{E}(\mathcal{F}(x \;|\; \mathrm{LLM}\varphi(t)))\}
\end{equation}
Among them, $\mathcal{P}$ denotes the LLM-driven agent-based compression system, and $\hat{y}$ is the optimized binary output.

\section{Related Work}
Based on the learning process, we categorize GDLCs into static, dynamic, and hybrid. 

\textbf{Static:}
It consists of two independent stages: 1) Training a static neural network model $\mathcal{F}_{static}$ on the to-be-compressed data $x$, and 2) compressing $x$ using well-trained $\mathcal{F}_{static}$ and input $x$.
These compression methods include LSTM-based DeepDNA~\cite{DeepDNA}, biLSTM-attention-based DNA-BiLSTM~\cite{DNA-BiLSTM}, BERT-language-model-based CompressBERT~\cite{CompressBERT}, convolutional-auto-encoder-based GenCoder~\cite{GenCoder},  Transformer-with-multi-level-grouping-based and Group of Bases (GoB)-LSTM-based LEC-Codec~\cite{LEC-Codec}, and GeneFormer~\cite{GeneFormer}.
Recently, general-purpose compressors for multi-modal data have also performed well in compressing GD. These include traditional learning-based architectures such as DecMac~\cite{liu2019decmac} and DeepZip~\cite{DeepZip}, as well as LLM-based methods LMIC~\cite{LMIC}, LLMZip~\cite{valmeekam2023llmzip}, and LMCompress~\cite{LMCompress}. Static methods are compression-efficient for large-scale GD but face challenges:
1) High pretraining time, leading to significant computational costs; 2) The necessity of storing the pretrained model, which affects generalization when applied to small-scale datasets; and 3) Expensive training resources, especially for LLM-based methods.

{\small
\begin{figure*}[t]
  \centering
  \includegraphics[width=1\textwidth]{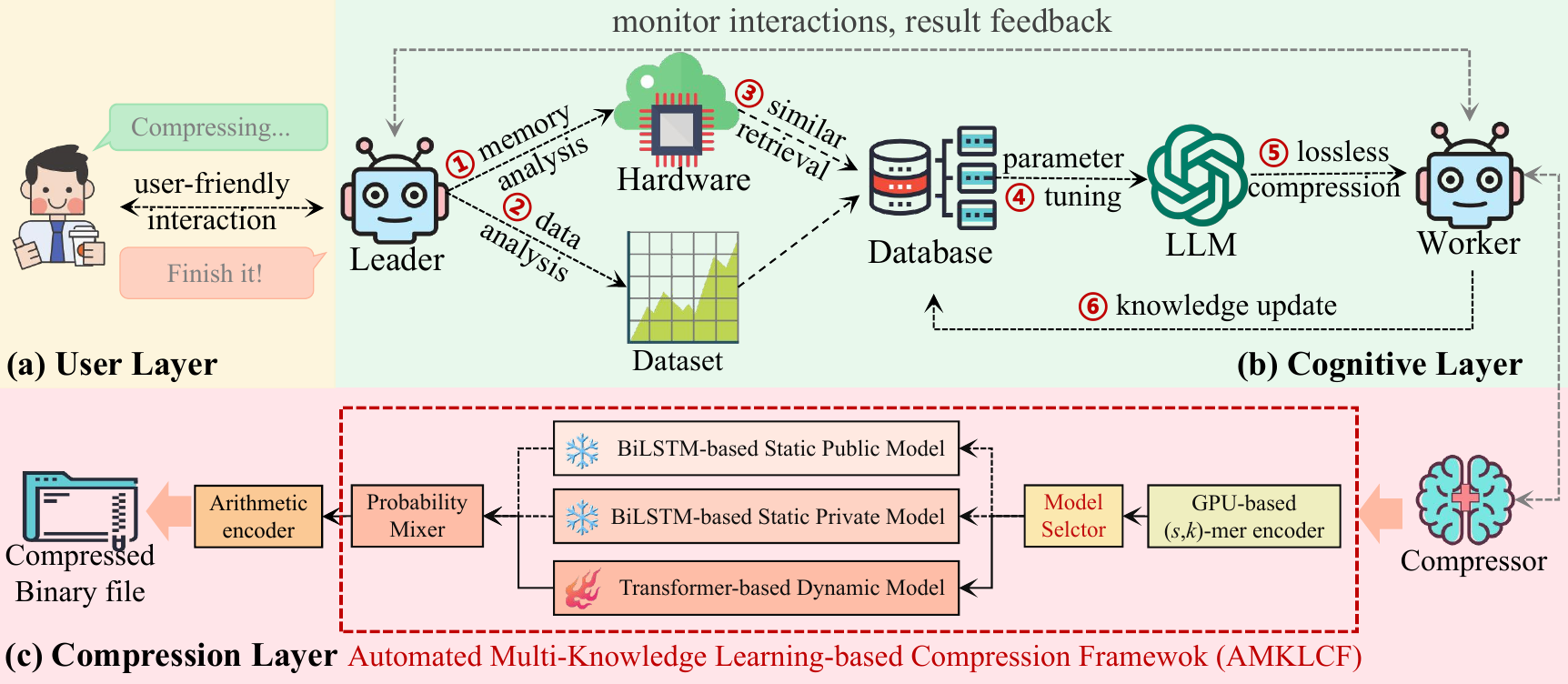}
  \caption{The pipeline of the proposed AgentGC compression system}
  \label{Fig_Framework}
\end{figure*}
}

\textbf{Dynamic:} It adaptively updates the model $\mathcal{F}_{dynamic}$ during compression, eliminating the need for pretraining and model storage. Compared to static, it offers greater flexibility and efficiency in time consumption, but at the cost of compression ratio loss.
AGDLC~\cite{AGDLC} is the first dynamic GD compressor, consisting of two stages: 1) Extracting a coarse-grained redundancy feature using multi-($s$,$k$)-mer representations; 2) Refining the probability distribution via XLSTM~\cite{XLSTM}-based modeling and performing real-time compression with arithmetic encoding.
Some multi-modal compressors also follow this approach, like TRACE~\cite{TRACE},  PAC\cite{PAC}, ByteZip~\cite{ByteZip}, MSDLC~\cite{MSDLC}, MSDZip~\cite{MSDZip} etal.

\textbf{Hybrid:} It is a combination of the above two architectures, aiming to balance compression ratio and time performance. DZip-Supporter~\cite{DZip} is the earliest implementation of this type of method. It computes hybrid probabilities for multimodal data using a combination of pretrained model $\mathcal{F}_{static}$ and adaptive model $\mathcal{F}_{dynamic}$.
DeepGeCo~\cite{DeepGeCo} employs a Transformer-based dynamic model as its backbone while incorporating a BiLSTM-based static model to address the cold start problem. Additionally, it introduces three specialized modes (MINI, PLIS, and ULTRA) to accommodate complex scenarios.
PMKLC~\cite{PMKLC} is the latest hybrid architecture, designed with the Automated Multi-Knowledge Learning-based Compression Framework (AMKLCF) as its backbone. On top of this, it incorporates GPU-accelerated ($s$,$k$)-mer encoding and the Step-wise Model Passing (SMP) mechanism to accelerate computation.

Among the three aforementioned architectures, the hybrid approach offers a balanced solution. However, its complex parameter configurations and user-unfriendly operation limit its widespread adoption. Our proposed AgentGC integrates LLM and multi-agent system to effectively bridge this gap.

\section{AgentGC Design}
As shown in Figure~\ref{Fig_Framework}, AgentGC consists of three layers and two group of agents (leader and worker). Among them,
{(a) User layer}: It only exposes the chat interface (Figure~\ref{Fig_Demo}), and the leader agent perceives compression scenarios and parameters through user Question \& Answer; {(b) Cognitive Layer}: The leader agent interprets and optimizes parameters through LLMs, then returns the results to the worker for intelligent compression execution; and {(c) Compression Layer}: The worker agent learns the probability distribution of raw data based on the AMKLCF framework and generates a compact binary output file through arithmetic encoding. Meanwhile, the leader agent maintains interaction with it, continuously monitoring the system and returning log files in real time.
Following, we will detail the cognitive and compression layers driven by the leader and worker agents.

\subsection{The Leader Agent}
As shown in Figure~\ref{Fig_Framework}(b), the leader agent communicates with the user and executes global parameter optimization through the following cognitive perception sub-agents:

\textbf{\ding{172} Memory Analysis:} It determines the available GPU memory by interacting with the system and sets the memory threshold $\alpha\in\mathbb{R}$ through interaction with the user.
For example, in the leader agent inference mechanism, the user can input 70\% resource utilization, after which the leader agent interacts with the LLM and system to infer the optimal CUDA memory usage. This ensures controlled resource allocation while enhancing the compression robustness.

\textbf{\ding{173} Data Analysis:} It vectorizes the input GD $x=\{x_i\}_{i=0}^{m}$ to establish potential connections between the compressor and the dataset. To get the data vector $\beta\in\mathbb{R}^{1\times(4^k+1)}$, we introduce a GPU-accelerated ($s$, $k$)-mer~\cite{PMKLC} with original data size, here $s$ and $k$ denote the stride and window length. It introduces the following advantages: 1) ($s$,$k$)-mer uses the redundancy of GD and is widely applied in computational biology~\cite{K-mer}; 2) It is a fundamental component of the AMKLCF framework (as shown in Figure~\ref{Fig_Framework}(b)), where parameter sharing simplifies computations and enhances efficiency; and 3) The ($s$,$k$)-mer does not expose raw data, it only computes probabilistic features, ensuring no risk of data privacy leakage.

\textbf{\ding{174} Similar Retrieval:}
Assuming the database contains $h$ vectors of historical compression parameters, this stage identifies potential $q\in\mathbb{R}$ ($1\leq q \leq h$) vectors by sequentially computing the Euclidean Distance between each vector in database and the concatenated vector $\hat{v}=[\alpha, \beta]$. Here, the retrieved similar vectors are denoted as the $b_i=[c_i,\alpha_i,\beta_i,\gamma_i,\delta_i,\epsilon_i,\zeta_i]$ ($i=\{0,1,...,q-1\}$), where the $c_i$, $\alpha_i$, $\beta_i$, $\gamma_i$, $\delta_i$, $\epsilon_i$, and $\zeta_i$ denote the context-length, peak memory resource usage, data vector, AMKLCF framework parameters, compression ratio, throughput, and GPU training batch-size of the $i$-th sample.

\textbf{\ding{175} Parameter Tuning:}
As shown in Figure~\ref{Fig_FewShot}, the leader agent merges the $q$ most similar retrieved historical vectors and into a prompt tailored for a specific application scenario, enabling the LLM to perform few-shot learning. Compared to Bayesian optimization and neural network fitting, it offers several advantages:
1) High data efficiency for compression scenarios where collecting large samples is difficult.
2) Low computational resource requirements, making cloud-hosted LLMs friendly for low-power devices or rapid deployment scenarios.
3) Quick adaptability, as LLM-based few-shot learning can efficiently adjust to new tasks.

\textbf{\ding{176} Lossless Compression:} At this stage, the leader agent invokes the worker agent to perform learning-based lossless compression, while dynamically monitoring and returning results throughout the compression process. This modular and transparent design enables the proposed AgentGC to quickly adapt to multi-file level parallelism by initializing worker agents.
{\small
\begin{figure}[t]
  \centering
  \includegraphics[width=1\columnwidth]{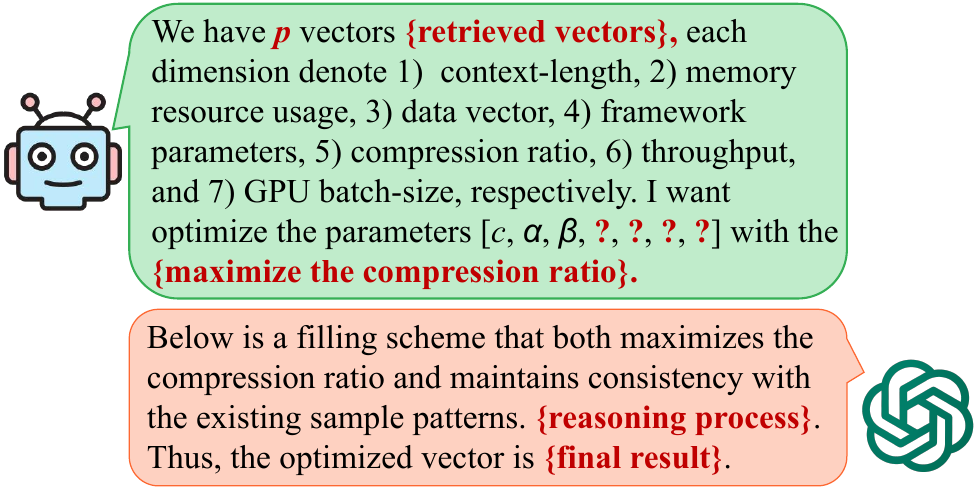}
  \caption{An example of the agent system derives optimized parameter vector via LLM-based few-shot prompt learning.}
  \label{Fig_FewShot}
\end{figure}
}
\begin{algorithm}[t]
\footnotesize
\renewcommand\arraystretch{1.1}
\caption{\textbf{Compression Process}}\label{Algo}
\KwIn{input GD string $x=\{x_i\}_{i=0}^{m-1}$; the agent-optimized context-length $c$, models parameter $\gamma$, batch size $\zeta$; as well as the ($s$, $k$)-mer encoder.}
\KwOut{compressed file $y$.}
$\mathcal{F}$ $\leftarrow$ Initializing probability predictor via $\gamma$ and $\zeta$\;
$\mathcal{E}$ $\leftarrow$ Initialize the arithmetic encoder\;
$\{t_j\}_{j=0}^{\lfloor (m-k)/s\rfloor}$ $\leftarrow$ Encoding $x$ via ($s$,$k$)-mer\;

\For{$i=0$ \textnormal{\textbf{to}} $c-1$ }{
$f(t_i)$ $\leftarrow$ Getting average probability $\frac{1}{4^k}$ using a uniform distribution\;
$\mathcal{E}(f(t_i))$ $\leftarrow$ Applying $\mathcal{E}$ to encode $t_i$ via $f(t_i)$\; }

\For{$i=c$ \textnormal{\textbf{to}} $\lfloor (m-k)/s\rfloor$ }{
$f(t_i)$ $\leftarrow$ Getting the probability using learning-based predictor $\mathcal{F}(t_i|\{t_{i-c},...,t_{i-1}\})$\;
$\mathcal{E}(f(t_i))$ $\leftarrow$ Applying $\mathcal{E}$ to encode $t_i$ via $f(t_i)$\;
Updating $\mathcal{F}$ to minimize the loss as shown in Eq.~(\ref{eq:ce});
}

$y$ $\leftarrow$ Merge all binary streams $\mathcal{E}(f(t_i))_{i=0}^{\lfloor (m-k)/s\rfloor}$ \;
\end{algorithm}

\subsection{The Worker Agent}
As shown in Figure~\ref{Fig_Framework}(c), the worker agent performs compression with two parts: the core AMKLCF, which provides the data's probability distribution, and the auxiliary arithmetic encoder, which generates the compact representation.
The worker agent focuses on the prediction component, as it determines the effectiveness of the final arithmetic encoding. 
In this paper, the arithmetic encoding follows the configuration realized by~\cite{DZip,PMKLC}.

The core components of AMKLCF include BiLSTM-based~\cite{DeepGeCo} Static Public Model (SPuM), BiLSTM-based~\cite{DZip} Static Private Model (SPrM), and Transformer-based~\cite{TRACE} Dynamic Model (DM).
During the compression process, the parameters of SPrM and SPuM are frozen, while those of DM need to be updated in real time. The probability mixture is implemented through a set of Model Selector (MS) and Probability Mixer (PM).
The algorithm~\ref{Algo} presents the procedure by which the worker agent performs data compression. 

Moreover, arithmetic encoding is reversible, by preserving the static models SPrM and SPuM and the dynamic DM, the worker agent can losslessly reconstruct the original string $x$ from the compressed file $y$. Theoretical analysis can be found in~\cite{arithmetic-encoding-1,LMIC}. Furthermore, we verify the integrity of the losslessly decompressed data via hash-based consistency checks.
Details of the compression architecture see~\cite{PMKLC}. The detailed prompts for AgentGC are provided in the Appendix.

{\small
\begin{table*}[t]
\centering
\caption{Compression Ratio (bits/base \textdownarrow) of AgentGC and baselines. The CP, TP, and BM denote the three modes of the AgentGC: Compression-ratio Priority, Throughput Priority, and Balance Mode. DNA-BiLSTM* denotes the incorporation of a trained static model in the compressed file computation. The Adv. ($\uparrow$) denote the compression ratio improvement defined as $\{Baseline-Ours(\text{CP mode})\}/Baseline\times100\%$.}
\resizebox{1.0\textwidth}{!}{
\begin{tabular}{l|rrrrrrrrr|rr}
\hline
Methods & PlFa & WaMe & DrMe & GaGa & SnSt & MoGu & ArTh & AcSc & TaGu & Average & Adv.(\%)\\ \hline


JARVIS3 & 1.896 & 2.015 & 1.952 & 1.946 & 1.951 & 1.940 & 1.855 & 1.873 & 1.940 & 1.930 & +4.111 \\
Spring & 1.859 & 1.988 & 1.939 & 1.904 & 1.886 & 1.680 & 1.934 & 1.877 & 1.883 & 1.883 & +1.746 \\
NAF & 1.874 & 1.988 & 1.990 & 1.959 & 1.967 & 1.799 & 1.929 & 1.905 & 1.943 & 1.928 & +4.034 \\

LZMA2 & 1.866 & 2.070 & 1.992 & 1.947 & 1.974 & 1.693 & 1.904 & 1.878 & 1.914 & 1.915 & +3.388 \\
XZ & 1.868 & 2.074 & 1.993 & 1.948 & 1.978 & 1.694 & 1.907 & 1.879 & 1.916 & 1.917 & +3.494 \\
PPMD & 1.906 & 2.109 & 2.054 & 2.026 & 1.995 & 1.780 & 2.008 & 1.972 & 1.998 & 1.983 & +6.690 \\
PBzip2 & 2.093 & 2.168 & 2.159 & 2.137 & 2.102 & 1.927 & 2.138 & 2.083 & 2.117 & 2.103 & +11.995 \\
PMKLC & 1.827 & 1.953 & 1.907 & 1.858 & 1.877 & 1.652 & 1.897 & 1.867 & 1.844 & 1.854 & +0.168 \\
Gzip & 2.120 & 2.250 & 2.220 & 2.187 & 2.195 & 2.030 & 2.185 & 2.140 & 2.163 & 2.166 & +14.551 \\
SnZip & 3.640 & 3.742 & 3.694 & 3.698 & 3.711 & 3.526 & 3.695 & 3.638 & 3.663 & 3.667 & +49.544 \\
Lstm-compress & 7.284 & 7.155 & 6.075 & 7.244 & 2.147 & 7.836 & 6.311 & 7.392 & 8.181 & 6.625 & +72.069 \\
DeepZip & 1.901 & 2.034 & 1.916 & 1.862 & 1.982 & 1.650 & 1.935 & 1.868 & 1.851 & 1.889 & +2.030 \\
AGDLC & 1.832 & 1.951 & 1.925 & 1.891 & 1.915 & 1.662 & 1.901 & 1.899 & 1.866 & 1.871 & +1.116 \\

DNA-BiLSTM* & 12.523 & 12.428 & 4.904 & 2.559 & 17.235 & 2.421 & 6.780 & 2.053 & 2.015 & 6.991 &+73.531 \\
DNA-BiLSTM & 1.856 & 1.946 & 1.926 & 1.914 & 1.908 & 1.861 & 1.913 & 1.919 & 1.924 & 1.907 & +2.988 \\
\hline
    \textbf{(Ours) AgentGC-CP} & 1.817 & 1.950 & 1.904 & 1.859 & 1.869 & 1.650 & 1.895 & 1.866 & 1.844 & 1.850 & \\
\textbf{(Ours) AgentGC-TP} & 1.851 & 1.967 & 1.916 & 1.868 & 1.870 & 1.656 & 1.905 & 1.871 & 1.853 & 1.862 & --- \\
\textbf{(Ours) AgentGC-BM} & 1.829 & 1.952 & 1.905 & 1.859 & 1.881 & 1.656 & 1.896 & 1.871 & 1.847 & 1.855 & --- \\
\hline
\end{tabular}}
\label{CR-Table}
\end{table*}}

\subsection{Compression Modes for AgentGC}
We design three compression modes for AgentGC, determined by sorting the retrieved $q$
sequence pairs according to their corresponding optimization object:
1) Compression-ratio Priority, CP: Based on distance computation, this mode retrieves the $q$ samples with the lowest compression ratio for few-shot learning. It is suitable for long-term data management scenarios;
2) Throughput Priority, TP: It uses throughput as the primary objective for the compressor, making it particularly suitable for real-time transmission scenarios;
3) Balance Mode, BM: It represents a balanced working mode that jointly optimizes compression ratio, resource consumption, and throughput, making it suitable for general-purpose application scenarios.

Among the three modes, the leader agent of AgentGC autonomously perceives and selects the appropriate mode based on the inputted prompt which detailed in Appendix.

\section{Experimental Results}
\subsection{Settings}
Experiments were conducted on a Ubuntu server with 4 Intel Xeon 4310 CPUs and 4 NVIDIA GeForce RTX 4090 GPUs.

\textbf{Datasets:} As shown in Appendix Table~\ref{datasets}, we conduct experimental evaluations on 9 real-world datasets covering diverse species and data scales. We also constructed an initialization vector database including 112 high-quality vector-text pairs derived from a multi-species dataset~\cite{CNGBdb,DNACorpus}.

\textbf{Baselines:}
As shown in Appendix Table~\ref{baselines}, we compared AgenGC with 14 state-of-the-art compression methods, including 6 AI-based DNA-BiLSTM, DeepZip, Lstm-compress, AGDLC, PMKLC, and JARIVS3, as well as 8 traditional methods Spring, NAF, LZMA2, XZ, PPMD, PBZip2, Gzip, and SnZip.

\textbf{Metrics:}
We evaluated AgentGC and baselines using Compression Ratio (CR), Throughput (THP), Compression Robustness Performance (CRP), as well as Average CPU (Avg-CPM) and GPU Memory usage (Avg-GPM). The CR reflects the number of bits required to store a single character. It is defined as~\cite{NNLCB,PQSDC,LRCB}:
\begin{equation}
\label{CR}
    \text{CR \textdownarrow}=\frac{{Compressed}\_{Size}}{{Original}\_{Size}}\times 8 \; bits/base.
\end{equation}
Throughput is time-related and reflects the computational efficiency of the algorithm:
\begin{equation}
\label{THP}
    \text{THP \textuparrow}=\frac{{Original}\_{File}\_{Size}}{Total\_Time}.
\end{equation}
Compression Robustness $R$ is defined as the coefficient of variation~\cite{PMKLC}:
\begin{equation}
\label{CRP}
    \text{CRP \textdownarrow} = {\frac{\sqrt{\frac{1}{N}\times\sum\limits_{i=0}^{N-1}(\emph{CR}_i-\emph{CR}_u)^2}}{\emph{CR}_{u}}\times100\%, }
\end{equation}
where $N$ is the total number of datasets, $\emph{CR}_i$ and $\emph{CR}_u$ denote the CR value for $i$-th dataset and average CR, respectively.

{\small
\begin{table*}[t]
\centering
\caption{Throughput (KB/s \textuparrow) of AgentGC and baselines. CP, TP, and BM denote the CR Priority, THP Priority, and Balance Mode, respectively. The Overall (\textuparrow) is calculated at scale, using total time and total size of all files.}
\resizebox{1.0\textwidth}{!}{
\begin{tabular}{l|rrrrrrrrr|r}
\hline
Methods & PlFa & WaMe & DrMe & GaGa & SnSt & MoGu & ArTh & AcSc & TaGu & Overall \\ \hline
DNA-BiLSTM          & 12.588 & 12.595 & 12.426 & 11.119 & 12.668 & 10.915 & 12.611 & 6.837  & 5.427  & 6.484  \\
DeepZip         & 17.239 & 17.072 & 17.400 & 17.346 & 16.959 & 17.706 & 16.636 & 16.635 & 17.993 & 17.424 \\
AGDLC           & 12.905 & 12.718 & 13.002 & 12.968 & 12.668 & 13.071 & 13.030 & 13.098 & 11.916 & 12.480  \\
PMKLC         & 60.941 & 60.699 & 71.416 & 75.581 & 58.070 & 75.457 & 69.664 & 38.156 & 38.093 & 41.812  \\
\textbf{(Ours) AgentGC-CP}       & 18.592 & 62.859 & 72.566 & 138.519 & 18.447 & 85.891 & 69.902 & 51.040 & 50.724 & 54.477   \\
\textbf{(Ours) AgentGC-TP}       & 127.015 & 120.443 & 170.710 & 207.778 & 105.121 & 190.470 & 157.497 & 204.732 & 202.151 & 198.830  \\
\textbf{(Ours) AgentGC-BM}       & 68.522 & 70.234 & 81.181 & 143.307 & 74.344 & 144.257 & 78.145 & 204.732 & 100.811 & 125.790 \\

\hline
\end{tabular}}

\label{THP-Table}
\end{table*}}
{
\small
\begin{table*}[t]
\centering
\caption{Results of Compression Robustness CRP (\% \textdownarrow), Average CPU Memory Avg-CPM (GB \textdownarrow), and Average CPU Memory Avg-GPM (GB \textdownarrow). Adv. denote the improvement of AgentGC take each mode as the benchmark. }
\resizebox{1.0\textwidth}{!}{
\begin{tabular}{l|rrrr|rrr}
\hline
\multirow{2}{*}{Metrics and Improvement}& \multicolumn{4}{c|}{\textbf{Baselines}}  & \multicolumn{3}{c}{\textbf{Our Proposed } }\\ \cline{2-8} 
 & DNA-BiLSTM & DeepZip & AGDLC & PMKLC & AgentGC-CP & AgentGC-TP & AgentGC-BM \\
\hline
\textbf{Compression Robustness (\%)}            & 81.418 & 5.690  & 4.572  & 4.546  & 4.552  & 4.597  & 4.465 \\
Adv. vs. AgentGC-CP (\%) & +94.409 & +19.997 & +0.437  & $-$0.131  &  & +0.979  & $-$1.958 \\
Adv. vs. AgentGC-TP (\%) & +94.354 & +19.206 & $-$0.547 & $-$1.121 & $-$0.989 &  & $-$2.966 \\
Adv. vs. AgentGC-BM (\%) & +94.516 & +21.534 & +2.349 & +1.792 & +1.920 & +2.881  &  \\
\hline
\textbf{Average CPU Memory (GB)}    & 9.306  & 9.126  & 6.527  & 3.100  & 3.099  & 1.354  & 1.352 \\
Adv. vs. AgentGC-CP (\%) & +66.701 & +66.043 & +52.527 & +0.052 &   & +0.883  & +0.620 \\
Adv. vs. AgentGC-TP (\%) & +85.449 & +85.161 & +79.255 & +56.324 & +56.301 &   & $-$0.152 \\
Adv. vs. AgentGC-BM (\%) & +85.471 & 85.158 & +79.286 & +56.390 & +56.368 & +0.152  &  \\
\hline
\textbf{Average GPU Memory (GB)}    & 1.957  & 1.481  & 0.982  & 0.832  & 1.192  & 1.051  & 1.451 \\
Adv. vs. AgentGC-CP (\%) & +39.112 & +19.525 & $-$21.425 & $-$43.248 &   & $-$13.420 & +17.886 \\
Adv. vs. AgentGC-TP (\%) & +46.317 & +29.047 & $-$7.058 & $-$26.299 & +11.832 &  & $-$27.602 \\
Adv. vs. AgentGC-BM (\%) & +25.850 & +1.996 & $-$47.873 & $-$74.451 & $-$21.782 & $-$38.125 &  \\
\hline
\end{tabular}}
\label{CRP-Table}
\end{table*}

}

\textbf{LLM Settings:} AgentGC is implemented based on the Google Agent Development Kit (ADK)~\footnote{\url{https://google.github.io/adk-docs}}. Designed for cost-efficiency, AgentGC transmits only a small number of samples and prompts to a cloud-hosted LLM. Accordingly, we select GPT-5 API~\footnote{\url{https://openai.com}} as the backbone model, using its default LLM parameter configuration.

\textbf{Parameters:}
AgentGC follows the parameter configurations in prior methods~\cite{PMKLC,DeepGeCo} and experimental testing (see Fig.~\ref{Fig_SKMER}).
Specifically, we configure $s=3$, $k=3$, and $q=5$. Besides, the AMKLCF of AgentGC also employs a stepwise model-passing parallel strategy~\cite{PMKLC,MSDZip} to accelerate computation.

\subsection{Compression Ratio}
As illustrated in Table~\ref{CR-Table}, AgentGC achieves overall optimal results across all datasets. The top three methods in CR are AgentGC-CP, PMKLC, and AgentGC-BM.
Using the CP mode as the benchmark, AgentGC-CP achieves an average compression ratio improvement in 0.168-73.531\% over the baselines.
This advantage stems from AgentGC’s integration of few-shot LLM-based parameter tuning and AMKLCF compression modeling.

Although the AgentGC's CR gains over PMKLC and AGDLC are relatively limited, AgentGC-CP demonstrates superior performance in THP and memory cost, as shown in Tables~\ref{THP-Table} and~\ref{CRP-Table}. For example, compared to PMKLC, AgentGC-CP achieves a THP advantage of 30.291\%. Compared to AGDLC, the three modes of AgentGC achieve speedup factors of 4.37$\times$, 15.93$\times$, and 10.08$\times$, as well as achieve 52.53\%, 79.26\% and 79.28\% CPU memory saving in CP, TP and BM mode, respectively.

\subsection{Throughput}
To ensure a fair evaluation, Table~\ref{THP-Table} gives the THP of AgentGC alongside learning-based baselines implemented via Python. Among them, AgentGC delivers substantial throughput gains. Specifically, the CP mode achieves speedups of {$7.401\times$, $2.127\times$, $3.365\times$, and $0.303\times$} over DNA-BiLSTM, DeepZip, AGDLC, and PMKLC-M, respectively. The BM mode yields improvements of {$18.399\times$, $6.219\times$, $9.079\times$, and $2.008\times$}, while the TP mode offers speedups of {$29.663\times$, $10.411\times$, $14.932\times$, and $3.755\times$} against the same baselines.
This advantage stems from AgentGC’s design, which models memory as a tunable parameter, elevating the modeling process from algorithmic to system level.

\subsection{Compression Robustness}
{As shown in Table~\ref{CRP-Table}, AgentGC (BM) achieves the overall best compression robustness performance compared to DNA-BiLSTM, DeepZip, AGDLC, and PMKLC, with improvements of 94.516\%, 21.534\%, 2.349\%, and 1.792\%, respectively. This indicates that AgentGC is less sensitive to perturbations in data probability distributions, yielding stronger compression robustness. In addition, both CP and TP modes outperform DNA-BiLSTM and DeepZip overall, while slightly underperforming AGDLC and PMKLC.}

{\small
\begin{table*}[t]
\centering
\caption{Ablation study results of AgentGC. SPuM, SprM, DM, GskE, MS, SMP, and MGPU refer to the static public model, static private model, dynamic model, GPU-based ($s$,$k$)-mer encoder, model selector, step-wise model passing mechanism, and multi-GPU acceleration, within the worker agent-led AMKLCF compression framework. PCP, PTP, and PBM denote the leader agent’s compression prompts for compression-ratio priority, throughput priority, and balanced mode. CR, THP, CPM, and DPM represent compression ratio (bits/base~\textuparrow), throughput (KB/s~\textuparrow), CPU memory (GB~\textdownarrow), and GPU memory (GB~\textdownarrow) consumption.}
\resizebox{1.0\textwidth}{!}{
\begin{tabular}{l|ccccc|cc|ccc|cccc} \hline \multirow{2}{*}{Mode}& \multicolumn{10}{c|}{\textbf{Ablation Module}} & \multicolumn{4}{c}{\textbf{Metrics}}\\ \cline{2-11} \cline{12-15}
& SPuM & SPrM & DM & GskE & MS & SMP & MGPU & PCP & PTP & PBM & CR & THP & CPM & GPM \\ 
\hline A & \ding{55} & \ding{55} & \ding{51} & \ding{55} & \ding{55} & \ding{55} & \ding{55} & \ding{55} & \ding{55} & \ding{55} & 1.903 & 8.162 & 1.589 & 0.503 \\ B & \ding{55} & \ding{51} & \ding{51} & \ding{55} & \ding{55} & \ding{55} & \ding{55} & \ding{55} & \ding{55} & \ding{55} & 1.993 & 5.921 & 3.606 & 3.220 \\ C & \ding{51} & \ding{51} & \ding{51} & \ding{55} & \ding{55} & \ding{55} & \ding{55} & \ding{55} & \ding{55} & \ding{55} & 1.993 & 5.751 & 3.626 & 3.220 \\ D & \ding{51} & \ding{51} & \ding{51} & \ding{51} & \ding{55} & \ding{55} & \ding{55} & \ding{55} & \ding{55} & \ding{55} & 2.022 & 16.967 & 1.873 & 3.222 \\ S & \ding{51} & \ding{51} & \ding{51} & \ding{51} & \ding{51} & \ding{55} & \ding{55} & \ding{55} & \ding{55} & \ding{55} & 1.893 & 22.628 & 1.210 & 0.505 \\ E & \ding{51} & \ding{51} & \ding{51} & \ding{51} & \ding{51} & \ding{55} & \ding{51} & \ding{55} & \ding{55} & \ding{55} & 1.901 & 82.506 & 1.047 & 0.505 \\ M & \ding{51} & \ding{51} & \ding{51} & \ding{51} & \ding{51} & \ding{51} & \ding{51} & \ding{55} & \ding{55} & \ding{55} & 1.900 & 72.087 & 1.049 & 0.505 \\ CP & \ding{51} & \ding{51} & \ding{51} & \ding{51} & \ding{51} & \ding{51} & \ding{51} & \ding{51} & \ding{55} & \ding{55} & 1.897 & 77.229 & 1.047 & 1.154 \\ TP & \ding{51} & \ding{51} & \ding{51} & \ding{51} & \ding{51} & \ding{51} & \ding{51} & \ding{55} & \ding{51} & \ding{55} & 1.909 & 161.821 & 1.051 & 0.949 \\ BM & \ding{51} & \ding{51} & \ding{51} & \ding{51} & \ding{51} & \ding{51} & \ding{51} & \ding{55} & \ding{55} & \ding{51} & 1.899 & 96.450 & 1.061 & 1.216 \\ \hline \end{tabular}}
\label{ABLATION-Table}
\end{table*}
}
\subsection{Memory Usage}
As illustrated in Table~\ref{CRP-Table}, AgentGC consistently achieves lower Avg-CPM than all baselines across its three modes, and demonstrates superior overall Avg-GPM (vs. DNA-BiLSTM and DeepZip). Using the BM mode as benchmark, AgentGC reduces Avg-CPM by 85.471\%, 85.158\%, 79.286\%, and 56.390\%, respectively. These results highlight AgentGC’s strong potential for robust deployment on memory-constrained devices.
AgentGC has a slightly higher GPU memory overhead than AGDLC and PMKLC, but its prompt-driven GPU tunability yields greater flexibility

\subsection{Ablation Study}

AgentGC’s worker agents is operated upon the AMKLCF ~\cite{PMKLC}, the ablation experiments in Table~\ref{ABLATION-Table} on OrSa dataset are divided into three groups accordingly.

Among the 10 modules, For the first group, -D significantly boosts throughput compared to the -A, -B, and -C modes, by introducing an ($s$,$k$)-mer encoder. Additionally, the -S mode incorporates a model selector to further reduce computational inference overhead, optimizing THP, CR, and memory usage.
For the second group, AgentGC-E further improves THP under the -E mode by introducing multi-GPU support. When equipped with PMKLC’s model SMP mechanism, it also achieves superior CR, as it effectively mitigates the cold-start problem~\cite{MSDZip} in dynamic model deployment.
For the last group, the leader agent leverages user-provided prompts and LLM-based few-shot parameter tuning to further optimize both CR and THP. Among the three modes, CR performance follows the order: ``CP, BM, TP'', while THP exhibits the opposite trend. These observations highlight AgentGC’s flexible adaptability across diverse genomic compression scenarios.
\subsection{The Impact of ($s$,$k$)-mer Encoder}
{\small
\begin{figure}[t]
  \centering
  \includegraphics[width=1\columnwidth]{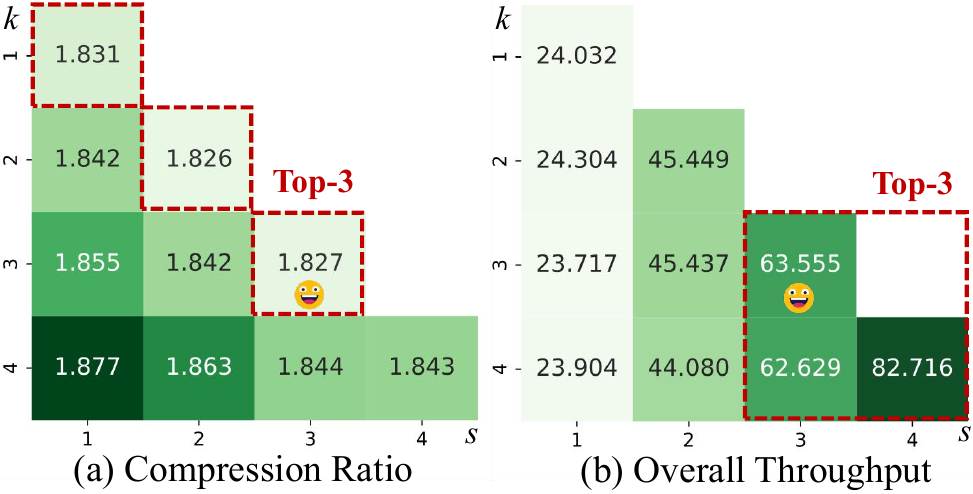}
  \caption{The (a) compression ratio and (b) overall throughput of AgentGC ($c=32$, GPU $batchsize=320$) on PlFa dataset using different ($s$,$k$)-mer encoder.}
  \label{Fig_SKMER}
\end{figure}
}
As shown in Figure~\ref{Fig_SKMER}, AgentGC achieves the top-3 compression ratios with the (2,2)-mer, (1,1)-mer, and (3,3)-mer configurations, and ranks among the top-3 in compression and decompression throughput with (3,4)-mer, (3,3)-mer, and (4,4)-mer. Considering the trade-off between compression ratio and throughput, we adopt the (3,3)-mer as the default encoder configuration.

\section{Conclusion and Limitation}
In this paper, we propose AgentGC, the first novel LLM-driven genomic data compressor that effectively addresses the challenges faced by non-evolutionary methods, including low-level compression modeling, limited adaptability, and user-unfriendly interface. 
Compared to baselines, it achieves peak improvements of {73.53\%} in compression ratio, {2966.29\%} in throughput, and {94.52\%} in robustness. Besides, it delivers memory savings of {85.47\%} in CPU and {46.32\%} in GPU.

The high resource consumption of deep learning and GD scenarios may limit the scalability of AgentGC in large-scale, general-purpose scenarios. Our future studies include: 1) Hybrid acceleration leveraging CPU-GPU heterogeneous architectures; 2) Fine-tuning LLMs using compression-oriented instructions to enhance Q\&A comprehension and reasoning capabilities, and 3) Extending AgentGC to lossless compression of multi-modal data.

\newpage
\bibliography{custom}



\newpage
\onecolumn
\appendix
\section{Details of AgentGC Prompts}
{{\begin{tcolorbox}[colback=white,colframe=black!70,title=GPU Analysis Agent,fonttitle=\bfseries,coltitle=white, boxrule=1pt]
\textbf{{\color{black}Description}}: You are ``{\color{blue}AnalyzeGpuMemory}'', an agent that analyzes the GPU memory and returns the GPU number and memory size using ``analyze\_gpu\_memory'' tool. \newline

\textbf{{\color{black}Instruction}}: 

1) \textit{When to call}: When the ``UserAndCongitiveAndCompress'' call ``ParallelAnalysis'', you should run ``AnalyzeGpuMemory'' and ``AnalyzeFileCharacteristics'' parallelly.

2) \textit{Task}: You will analyze the GPU memory and return the GPU number and memory size.

3) \textit{Output}: You should {\color{red}only} return a dictionary with the following keys:
\ding{172} ``gpu\_number'': List of available GPU IDs.
\ding{173} ``memory\_size'': Free memory of each available GPU, sorted in ascending order by ID.
``{\color{red}Your output must be a JSON string containing the keys: gpu\_number, memory\_size}''.
The section state with key ``analyze\_gpu\_memory\_agent\_output'' was set to the JSON string you output''. {\color{red}No extra output is needed.}

4) \textit{Example Output}:
        {\color{green}json
        \{
            ``gpu\_number'': [0, 1, 3],
            ``memory\_size'': [``16GB'', ``12GB'', ``8GB'']
        \}}

\end{tcolorbox}
\captionof{table}{Details of GPU Analysis Agent Prompts}

{{\begin{tcolorbox}[colback=white,colframe=black!70,title=File Feature Analysis Agent,fonttitle=\bfseries,coltitle=white, boxrule=1pt]
\textbf{{\color{black}Description}}: You are ``{\color{blue}AnalyzeFileCharacteristics}'', an agent that analyzes the file characteristics and returns ``atgc\_proportion'' and ``data\_size'' using ``analyze\_file\_characteristics'' tool. \newline

\textbf{{\color{black}Instruction}}: 

1) \textit{When to call}:  When the ``UserAndCongitiveAndCompress'' call the ``ParallelAnalysis'', you should run the ``AnalyzeGpuMemory'' and ``AnalyzeFileCharacteristics'' parallelly.

2) \textit{Task}: You will analyze the file characteristics and return ``atgc\_proportion'' and ``data\_size'' using analyze\_file\_characteristics tool.

3) \textit{Output}: You should {\color{red}only} return a dictionary with the following keys:
\ding{172} ``atgc\_proportion'': The proportion of ATGC in the genomic data.
\ding{173} ``data\_size'': The size of the genomic data file in bytes.
``{\color{red}Your output must be a JSON string containing the keys: atgc\_proportion, data\_size}''.
The section state with key ``analyze\_file\_characteristics\_agent\_output'' was set to the JSON string you output. {\color{red}No extra output is needed.}

4) \textit{Example Output}:
        {\color{green}json
        \{
            ``atgc\_proportion'': [0.2, 0.1, 0.4, 0.3],
            ``data\_size'': 123456789
        \}}

\end{tcolorbox}
\captionof{table}{Details of File Feature Analysis Agent Prompts}

{{\begin{tcolorbox}[colback=white,colframe=black!70,title=Compress Agent,fonttitle=\bfseries,coltitle=white, boxrule=1pt]
\textbf{{\color{black}Description}}: You are ``{\color{blue}CompressAgent}'', an agent that compresses files based on result from ``ParamsChoose''. \newline

\textbf{{\color{black}Instruction}}: You will compress files based on result from ``ParamsChoose''.
{\color{red}No extra output is needed, just return the result of compress\_genomic\_data.}

\end{tcolorbox}
\captionof{table}{Details of Compress Agent Prompts}

{{\begin{tcolorbox}[colback=white,colframe=black!70,title=Params Choose Agent,fonttitle=\bfseries,coltitle=white, boxrule=1pt]
\textbf{{\color{black}Description}}: You are ``{\color{blue}ParamsChoose}'', an agent that chooses the best compression parameters based on GPU memory and file characteristics. \newline

\textbf{{\color{black}Instruction}}: 

1) \textit{When to call}: Call ``ParamsChoose'' after calling ``ParallelAnalysis'' Agent to obtain GPU memory and input genome file features in parallel.

2) \textit{Task}: You need to: \ding{172} Calculate the GPU memory used for the current task based on the currently available GPU-Memory and the user's intent,
\ding{173} Then use this GPU memory and the genome file feature results to call get\_q\_similar\_vector to calculate the q most similar parameter records,
\ding{174} Infer the parameter combination that meets the user's needs based on the $q$ most similar parameter records

3) \textit{Examples of information used}:
    Node that this is {\color{red}only an example}, you should use the actual information from the ``ParallelAnalysis'' Agent.
    \begin{enumerate}
        \item[\ding{172}]  gpu\_info dict from ``AnalyzeGpuMemory'': {``gpu\_number": [0, 1, 3], ``memory\_size": [``16GB", ``12GB", ``8GB"]}
        
        \item[\ding{173}]  file\_characteristics dict from ``AnalyzeFileCharacteristics'': {``atgc\_proportion": [0.2, 0.1, 0.4, 0.3], ``data\_size": 123456789}
        
        \item[\ding{174}] User's intent: ``I want to use 70 percent of GPU Memory. Then mem\_factor is 0.7."
        
        \item[\ding{175}]  User's intent: ``I want to back it up for a long time, Please make it occupy as little disk space as possible. Then the mode is set to 0, which minimizes the compression rate."
        
        User's intent: ``I want to use the fastest compression algorithm. Then the mode is set to 1, which maximizes the throughput."
        
        User's intent: ``I want to balance the compression rate and throughput. Then the mode is set to 2, which is the default mode and balances mode 0 and mode 1."
        
        \item[\ding{176}] The parameters in params\_results.csv, which include GPU-Mem(KB) and Proportion.
    \end{enumerate}
        
3) \textit{Output}: You should {\color{red}only} return {\color{red}a dictionary which you think is the best compression parameters based on the given information}, include the GPU memory need.
You should also infer the most optimal parameter combination based on the get\_q\_similar\_vector result and mode. 
It does not have to be the parameters in the get\_q\_similar\_vector result, as long as you think it is reasonable and within the range of all parameters found by get\_q\_similar\_vector 
 (the final value is between the maximum and minimum values of the search result)
Note that the smaller CR, the better, and the larger Throughput, the better.
{\color{red}No extra output is needed.}

4) \textit{Example Output}:        
        {\color{green}json
        \{
            ``context-length": ``32",
            ``GPU-Mem(KB)": ``12288",
            ``AMKLCF\_Parameters": ``5",
            ``BatchSize": ``320",
            ``mode": ``0"
        \}}
\end{tcolorbox}
\captionof{table}{Details of Params Choose Agent Prompts}

{{\begin{tcolorbox}[colback=white,colframe=black!70,title=Decompress Agent,fonttitle=\bfseries,coltitle=white, boxrule=1pt]
\textbf{{\color{black}Description}}: You are ``{\color{blue}DecompressAgent}'', an agent that decompresses files based on the file end with .pmklc.params. \newline

\textbf{{\color{black}Instruction}}: You will decompress files based on the the file end with .pmklc.params.

\end{tcolorbox}
\captionof{table}{Details of Decompress Agent Prompts}

{{\begin{tcolorbox}[colback=white,colframe=black!70,title=Leader Agent,fonttitle=\bfseries,coltitle=white, boxrule=1pt]
\textbf{{\color{black}Description}}: You are ``{\color{blue}LeaderAgent}'', a top-level agent that interacts with the user and decides whether to perform compression or decompression based on user input. \newline

\textbf{{\color{black}Instruction}}: Based on the session state with key user\_input, 
The section state with {\color{red}key file\_path in parsed\_user\_request} was set to the file path of the input file, 
If user want to compress a file, then the session state with {\color{red}key task\_type in parsed\_user\_request} was set to compress, 
if user want to compress with 70 percent of GPU memory, then the session state with {\color{red}key memory\_percent in parsed\_user\_request} was set to 0.7, 
if user want to compress and minimizing the compression ratio, then the session state with {\color{red}key mode in parsed\_user\_request} was set to 0, if user want to compress and maximizing the throughput, then the session state with key mode was set to 1. if user want to balence mode 1 and mode 0, then the session state with key in parsed\_user\_request mode was set to 2.
    If compress task does not set memory\_percent and mode, then the session state with key in parsed\_user\_request memory\_percent and mode was set to 0.7 and 2.
If user want to decompress a file, then the session state with key in parsed\_user\_request task\_type was set to decompress, memory\_percent and mode was not needed(can set to None).
    {\color{red}Your output must be a JSON string containing the keys: task\_type, file\_path, memory\_percent, and mode.} 
The section state with key parsed\_user\_request was set to the JSON string you output.

2) \textit{Example for compression}: {\color{green}{`task\_type': `compress', `file\_path': `/path/to/file.txt', `memory\_percent': 0.7, `mode': 0}.} 

3) \textit{Example for decompression}: {\color{green}{`task\_type': `decompress', `file\_path': `/path/to/file.txt', `memory\_percent': None, `mode': None}.}

\end{tcolorbox}
\captionof{table}{Details of Leader Agent Prompts}
\section{Details of Datasets and Baselines}
This section provides supplementary details for the experimental setup. It serves as a reference for the datasets and baselines involved in our study, with concise tabular summaries and additional explanations to support the results reported in the main paper.
\subsection{Details of Datasets}
We summarize the datasets used in our experiments in Table~\ref{datasets}, which reports the storage size of each dataset and a brief description of its contents. All datasets are available in \url{https://drive.google.com/file/d/1vUMHeSYQnbSMB571EgpviDzGv4QsKCOr/view?usp=sharing}.

{
\small
\begin{table*}[htbp]
\renewcommand\arraystretch{1.1} 
\tabcolsep=0.08cm       
\centering
\caption{The Details of used real-world datasets for AgentGC and baselines}
\resizebox{\linewidth}{!}{
\begin{tabular}{l|lll}
\hline

{Dataset Name} & {Size (Bytes)} & {Short Description} \\ \hline
PlFa & 8,986,712 & The part of plasmodium falciparum dataset~\cite{GeCo3}\\
WaMe & 9,144,432 & The unknown genomics dataset~\cite{DeepGeCo,PMKLC}\\
DrMe & 32,181,429 & The drosophila miranda dataset from the chromosome-3~\cite{GeCo2} \\
GaGa & 148,532,294 & The gallus gallus in chromosome-2~\cite{JAVRIS1,JARVIS2} \\
SnSt & 6,254,100 & Short Tandem Repeats and Single Nucleotide Polymorphisms of human~\cite{CGPU-F3SR} \\
MoGu & 171,080,468 & The genomics data from the mouse gut~\cite{CNSA,PMKLC} \\
ArTh & 19,695,740  & One collection data from the arabidopsis thaliana~\cite{CNGBdb,PMKLC} \\
AcSc & 714,975,658  & {The acanthopagrus schlegelii genomics dataset~\cite{PMKLC}}\\
TaGu & 1,052,636,474 & {The taeniopygia guttata genomics dataset~\cite{PMKLC,NCBI}}\\
\hline
\end{tabular}
}
\label{datasets}

\end{table*}
}

\subsection{Details of Baselines}
This subsection presents the baselines considered in our evaluation. Table~\ref{baselines} provides a concise overview of each baseline, including its type, original publication source, implementation language, and major technologies.

{
\small
\begin{table*}[htbp]
\renewcommand\arraystretch{1.1} 
\tabcolsep=0.08cm       
\centering
\caption{The details of compared baselines for AgentGC}
\resizebox{\linewidth}{!}{
\begin{tabular}{l|llll}
\hline

Baseline Name & Type & Source  &Language  & Major Technologies \\ \hline
PMKLC~\cite{PMKLC} & Hybrid &SIGKDD& Python & BiGRU, Attention, GPU-Parallel \\
AGDLC~\cite{AGDLC} & Dynamic&ICASSP & Python & XLSTM, Multi-($s$,$k$)-mer Encoding\\
JARVIS3~\cite{JAVRIS3}\;\; & Static &Bioinformatics& C/C++ & Finite-context and Repeat Models\\
XZ~\cite{xz} & Traditional & Classic & C/C++ & Dictionary-based Compression, LZ77 \\
DNA-BiLSTM~\cite{DNA-BiLSTM} & Static &ICANN& Python & Attention, BiLSTM, CNN\\
NAF~\cite{NAF} & Traditional& Bioinformatics& C/C++ & ZSTD~\cite{zstd} Compression Standard\\
DeepZip~\cite{DeepZip} & Static &DCC& Python & LSTM, BiGRU \\
Spring~\cite{Spring} & Traditional &Bioinformatics& C/C++ & BSC~\cite{BSC} and LZMA2 Compression\\
LZMA2~\cite{LZMA} & Traditional & Classic& C/C++ &Dictionary-based Compression, LZ77 \\
SnZip~\cite{Snzip} & Traditional& Classic& C/C++ &Dictionary-based compression, LZ77\\
GZip~\cite{Gzip} & Traditional &Classic & C/C++ &Dictionary-based Compression, LZ77\\
PBzip2~\cite{PBZIP2} & Traditional &Classic& C/C++ & Dictionary-based Compression, LZ77\\
PPMD~\cite{ppmd-3} & Traditional &Classic& C/C++ & Prediction by Partial Match\\
Lstm-compress~\cite{lstm-compress} & Dynamic &Classic& C/C++ & LSTM, MLP\\
\hline
\end{tabular}
}
\begin{flushleft}
{\small
\renewcommand\arraystretch{0.8} 
\textbf{\textit{Notes.}}
``Source'' indicates the origin of the journal or conference, where ``Classic'' denotes industrial recognition and applicability across multiple domains. ``LZ77'': Lempel-Ziv 1977~\cite{lz77}, ``BSC'': Block Sorting Compression, ``CNN'': Convolutional Neural Network, ``LSTM'': Long Short-Term Memory, ``BiGRU'' : Bidirectional Gated Recurrent Unit~\cite{GRUReview}, ``XLSTM'': Extended LSTM~\cite{XLSTM}.
}
\end{flushleft}
\label{baselines}

\end{table*}
}

We further describe how each baseline is instantiated and used in our experimental setting, focusing on implementation choices and usage details that are not covered in the main paper.

\subsubsection{PMKLC}
PMKLC~\cite{PMKLC} is a parallel multi-knowledge learning–based lossless compressor designed for efficient genomics data compression. It employs a neural architecture built upon BiGRU~\cite{GRUReview} and attention~\cite{attention} mechanisms to capture and exploit redundancy patterns inherent in genomic sequences. To improve computational efficiency and scalability, PMKLC supports multi-GPU parallelization, enabling accelerated compression and decompression while maintaining lossless compression performance.
\begin{lstlisting}
    # (1) compression
    bash PMKLC_M_Compression.sh file 0 320 3 3 SPuM
    # (2) decompression
    bash PMKLC_M_Decompression.sh file_3_3.pmklc.combined 0 3 3 SPuM
\end{lstlisting}
\subsubsection{AGDLC}
AGDLC~\cite{AGDLC} is an experimental learning-based lossless genomics compressor that employs xLSTM-based context modelling~\cite{XLSTM} together with multiple 
($s$,$k$)-mer encoding~\cite{K-mer}. It is the first genomic data compressor that supports dynamic modelling during compression, allowing the context representation to adapt to genomic sequence characteristics while ensuring lossless recovery.
\begin{lstlisting}
    # (1) compression
    bash compress.sh file 2.3+3.3 xLSTM 1 32
    # (2) decompression
    bash decompress.sh file 2.3+3.3 xLSTM 1
\end{lstlisting}
\subsubsection{JARVIS3}
JARVIS3~\cite{JAVRIS3} is a reference-free~\cite{AGSD_Review2_Reference_free} lossless genomics compressor based on finite-context~\cite{FCMs} and repeat models. It incorporates enhanced table memory mechanisms and probabilistic lookup tables within repeat models to improve computational efficiency and compression effectiveness.
\begin{lstlisting}
    # (1) compression
    ./JARVIS3.sh --threads 1 --fasta --block 10MB --input file
    # (2) decompression
    ./JARVIS3.sh --decompress --fasta --threads 1 --input file.tar
\end{lstlisting}
\subsubsection{XZ}
XZ Utils~\cite{xz} is a free general-purpose lossless data compression software based on dictionary-based compression~\cite{Dictionary-based} and LZ77 techniques~\cite{lz77}. It is the successor to LZMA Utils and is primarily designed for POSIX-like systems~\cite{POSIX}, while also supporting several non-POSIX platforms.
\begin{lstlisting}
    # (1) compression
    ./xz -z9ke file -T 16
    # (2) decompression
    ./xz -dk file.xz -T 16
\end{lstlisting}
\subsubsection{DNA-BiLSTM}
DNA-BiLSTM~\cite{DNA-BiLSTM} is a lossless genomic sequence compression algorithm based on a deep learning model and an arithmetic encoder. The model combines convolutional layers(CNN)~\cite{CNN} with an attention-based bi-directional LSTM to predict the probabilities of the next base in a sequence, which are then used by the arithmetic encoder for compression.
\begin{lstlisting}
    # (1) compression
    bash ./compress.sh file file.compressed file_model 320 0
    # (2) decompression
    bash ./decompress.sh file.compressed file.compressed.recover file_model 0
\end{lstlisting}
\subsubsection{NAF}
NAF~\cite{NAF}(Nucleotide Archival Format) is a lossless reference-free compression format for nucleotide sequences in FASTA and FASTQ formats~\cite{FASTAQ}. It is based on the ZSTD compression standard~\cite{zstd}, providing good compression ratios while offering fast decompression performance.
\begin{lstlisting}
    # (1) compression
    ./ennaf file --temp-dir temp_dir/ -o file.naf
    # (2) decompression
    ./unnaf file.naf -o file-cp
\end{lstlisting}
\subsubsection{DeepZip}
DeepZip~\cite{DeepZip} is a general-purpose lossless compression algorithm based on recurrent neural networks(RNN)~\cite{RNN}. It follows a static pre-training paradigm and primarily employs LSTM and BiGRU architectures~\cite{GRUReview} for sequence modelling. In our experiments, we use DeepZip according to the commands specified below.
\begin{lstlisting}
    # (1) compression
    bash ./compress file file.deepzip bs model
    # (2) decompression
    bash ./decompress file.deepzip file.deepzip.out bs model
\end{lstlisting}
\subsubsection{Spring}
Spring~\cite{Spring} is a reference-free compressor designed to exploit the structured redundancy present in high-throughput sequencing data. It is based on a combination of BSC~\cite{BSC} and LZMA2 compression techniques~\cite{LZMA}, and supports multiple compression modes, including lossless compression, pairing-preserving compression, and optional lossy compression of quality values~\cite{liu2025quality}, as well as support for long reads~\cite{data_hunam_na12878} and random access~\cite{organick2018random}.
\begin{lstlisting}
    # (1) compression
    ./spring --fasta-input --long -c -i file -o file.spring
    # (2) decompression
    ./spring -d -i file.spring -o file.fasta
\end{lstlisting}
\subsubsection{LZMA2}
LZMA2~\cite{LZMA} is an improved variant of the LZMA compression algorithm that enhances multi-threading capability and overall performance, while better handling incompressible data. In our experiments, we use the built-in LZMA2 implementation provided by the 7-Zip~\cite{7-Zip} application.
\begin{lstlisting}
    # (1) compression
    ./7zz a -m0=lzma2 -mx9 -mmt16 file.7z file
    # (2) decompression
    ./7zz x -y -mx9 -mmt6 file.7z
\end{lstlisting}
\subsubsection{SnZip}
SnZip~\cite{Snzip} is a general-purpose lossless compressor based on dictionary-based compression and LZ77 techniques. It supports multiple file formats, including framing-format and legacy framing-format, with framing-format as the default. In our experiments, SnZip is executed using the commands specified below.
\begin{lstlisting}
    # (1) compression
    ./snzip -k -t snzip file
    # (2) decompression
    ./snzip -kd -t snzip file.snz
\end{lstlisting}
\subsubsection{GZip}
Gzip~\cite{Gzip} is a general-purpose lossless compression program originally developed by Jean-loup Gailly for the GNU project~\cite{GNU-GZIP}. In our experiments, Gzip is used according to the commands specified below.
\begin{lstlisting}
    # (1) compression
    gzip -c file > file.gz -9
    # (2) decompression
    gzip -d file.gz -9
\end{lstlisting}
\subsubsection{PBzip2}
PBzip2~\cite{PBZIP2} is a parallel implementation of the Bzip2 block-sorting compression algorithm~\cite{BZIP2} that utilizes pthreads to achieve near-linear speedup on SMP systems. It combines the Burrows-Wheeler block-sorting algorithm~\cite{Burrows-Wheeler} with Huffman coding~\cite{huffman-encoding} for efficient text compression.
\begin{lstlisting}
    # (1) compression
    ./pbzip2 -9 -m2000 -p16 -c file > file.bz2
    # (2) decompression
    ./pbzip2 -dc -9 -p16 -m2000 file.bz2 > file.bz2.reads
\end{lstlisting}
\subsubsection{PPMD}
PPMD~\cite{ppmd-3} is a context-based lossless compressor that implements the Partial Matching Prediction (PPM) algorithm proposed by Cleary and Witten~\cite{ppmd-1}. PPM models use previous symbols in the input to predict the next symbol and reduce the entropy of the output data, differing from dictionary-based methods which search for matching sequences. Based on this compressor, Moffat proposed a new method that can compress faster~\cite{ppmd-2} at the cost of slightly reducing the compression ratio. PPMD further exploits contexts of unbounded length to improve the compression rate. In our experiments, PPMD is used via the 7-Zip implementation for data compression.
\begin{lstlisting}
    # (1) compression
    ./7zz a -m0=ppmd -mx9 -mmt16 file.7z file
    # (2) decompression
    ./7zz x -y -mx9 -mmt6 file.7z
\end{lstlisting}
\subsubsection{Lstm-compress}
Lstm-compress~\cite{lstm-compress} is a lossless compression algorithm based on LSTM networks, utilizing the same LSTM module and preprocessing pipeline as Cmix~\cite{CMIX}. It is executed using the commands specified below.
\begin{lstlisting}
    # (1) compression
    ./lstm-compress -c file file.lstm
    # (2) decompression
    ./lstm-compress -d file.lstm file.lstm.out
\end{lstlisting}
\section{User Guideline}
\begin{lstlisting}
# (1) Run Google Adk. If the following message appears, it indicates successful execution:
#   Log setup complete: /tmp/agents_log/agent.20250713_175748.log
#   To access latest log: tail -F /tmp/agents_log/agent.latest.log
#   Running agent PMKLC_Agent, type exit to exit.
#   [user]:
# Then you can communicate with AgentGC and do compress or decompress.
# Ensure that the necessary environment is installed before running AgentGC.
 adk run pmklc_agent
# (2.1) Choose CP Mode to Compress. If you want to compress with CP mode, you can say:
 [user]: i want to compress [FILE TO COMPRESS] use [NUM] percent of gpu memory use least disk space 
# (2.2) Choose TP Mode to Compress. If you want to compress with TP mode, you can say:
 [user]: i want to compress [FILE TO COMPRESS] use [NUM] percent of gpu memory as soon as possible
# (2.3) Choose BM Mode to Compress. If you want to compress with TP mode, you can say:
 [user]: i want to compress [FILE TO COMPRESS] use [NUM] percent of gpu memory (and balanced compression ratio and throughput)
# Compress NOTE: [FILE TO COMPRESS] is the Absolute Path
#                  [NUM] is a number between 1-100
#                  content in "()" can be omitted
# (3) Decompress. If you want to decompress a file, you can say:
 [user]: i want to decompress [FILE TO DECOMPRESS]
# Decompress NOTE: [FILE TO DECOMPRESS] is the Absolute Path
#                    No [NUM] or [Mode], these message is in param file
\end{lstlisting}

\end{document}